\documentclass[runningheads]{llncs}
\usepackage{graphicx}

\usepackage{tikz}
\usepackage{comment}
\usepackage{amsmath,amssymb} 
\usepackage{color}
\usepackage{booktabs}
\usepackage{multirow}
\usepackage{booktabs}
\usepackage{enumitem}
\usepackage{amsfonts}
\usepackage{fancyhdr}
\usepackage[accsupp]{axessibility}  

\begin{document}
\pagestyle{headings}
\mainmatter
\def\ECCVSubNumber{7215}  

\title{Bi-PointFlowNet: Bidirectional Learning for Point Cloud Based Scene Flow Estimation} 

\titlerunning{Bidirectional Learning for Point Cloud Based Scene Flow Estimation}
%
\author{Wencan Cheng\inst{1}\orcidID{0000-0002-7996-0236} \and
Jong Hwan Ko\inst{2}\orcidID{0000-0003-4434-4318}}
\authorrunning{W. Cheng, J. H. Ko.}
%
\institute{Department of Artificial Intelligence, Sungkyunkwan University, Suwon 16419, South Korea \and
College of Information and Communication Engineering, Sungkyunkwan University, Suwon 16419, South Korea\\
\email{\{cwc1260,jhko\}@skku.edu}}
\maketitle
\thispagestyle{fancy}
\lhead{\scriptsize{Accepted as a conference paper at European Conference on Computer Vision (ECCV) 2022}}

\begin{abstract}
Scene flow estimation, which extracts point-wise motion between scenes, is becoming a crucial task in many computer vision tasks. However, all of the existing estimation methods utilize only the unidirectional features, restricting the accuracy and generality. This paper presents a novel scene flow estimation architecture using bidirectional flow embedding layers. The proposed bidirectional layer learns features along both forward and backward directions, enhancing the estimation performance.  
In addition, hierarchical feature extraction and warping improve the performance and reduce computational overhead. Experimental results show that the proposed architecture achieved a new state-of-the-art record by outperforming other approaches with large margin in both FlyingThings3D and KITTI benchmarks. Codes are available at \url{https://github.com/cwc1260/BiFlow}.

\keywords{Scene flow estimation, Point cloud, Bidirectional learning }
\end{abstract}

\section{Introduction}
\label{sec:intro}

A scene flow estimation task is to capture the point-wise motion from two consecutive frames. As it provides the fundamental low-level information of dynamic scenes, it has become an essential step in various high-level computer vision tasks including object detection and motion segmentation. Therefore, accurate scene flow estimation is crucial for perceiving dynamic environment in real-world applications such as autonomous driving and robot navigation \cite{wang2021hierarchical,kittenplon2021flowstep3d}.

\begin{figure}
\centering
\includegraphics[width=.5\linewidth]{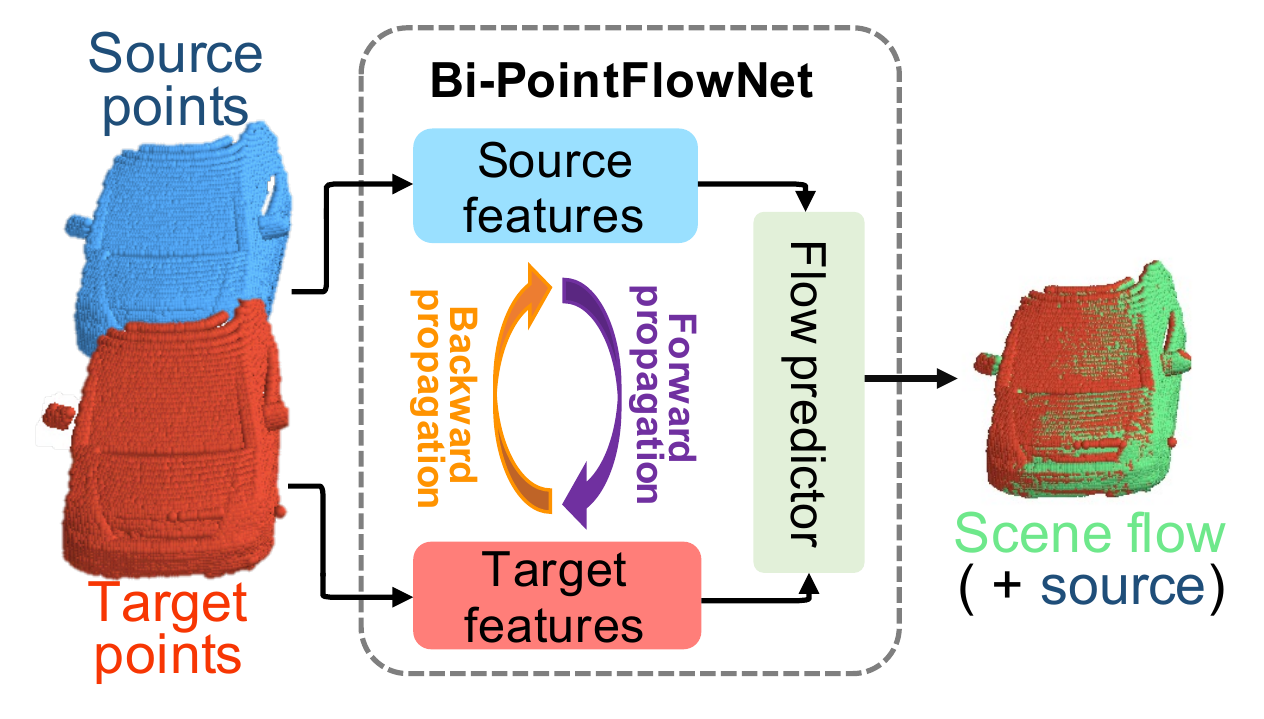}
\caption{Illustration of the bidirectional learning for scene flow estimation. The features extracted from each input frame are propagated bidirectionally for generating augmented feature representations that benefit scene flow estimation. The estimated scene flows are warped with the source frame for a clear comparison with the target frame.
}
\label{fig:concept}
\end{figure}

Early scene flow estimation approaches employed RGB images as an input. However, due to the increasing applications of LiDAR sensors that can capture dynamic scenes in the form of three-dimensional (3D) point clouds, scene flow estimation using point cloud has been actively studied. FlowNet3D \cite{liu2019flownet3d} proposed the first point cloud based estimation model using the hierarchical architecture of PointNet++ \cite{qi2017pointnet++}. 
Based on this scheme, several studies \cite{tishchenko2020self,gu2019hplflownet} proposed the multi-scale correlation propagation structure for more accurate estimation. Recently, PointPWC \cite{wu2020pointpwc} significantly improved the estimation performance using regression of multi-scale flows in a coarse-to-fine manner. Another study \cite{puy2020flot} proposed integration of an optimal transport-solving module in a neural network for estimating scene flow. 

All of these existing methods utilized only the unidirectional feature propagation (i.e., propagating source point features to target points) for calculating flow correlations. Meanwhile, various models for natural language processing (NLP) tasks \cite{schuster1997bidirectional,graves2005framewise,devlin2018bert,lee2020biobert} showed that the bidirectionally-learned features can significantly improve the performance due to their strong contextual information. 

Since scene flow estimation is also a temporal sequence processing task, bidirectional learning can boost the estimation performance. Bidirectional configuration has already proved its effectiveness on optical flow estimation, which is similar representation as scene flow estimation \cite{hur2017mirrorflow,wang2018occlusion,janai2018unsupervised,hur2019iterative,liu2019selflow}. However, to the best of our knowledge, there is no prior work that utilized bidirectional learning for the estimation of scene flow in the 3D space.

Based on this motivation, we propose Bi-PointFlowNet, a novel bidirectional architecture for point cloud based scene flow estimation.
As shown in Fig. \ref{fig:concept}, the bidirectional correlations can be learned by forward-propagation from the source features and backward-propagation from the target features. 
Therefore, each frame contains knowledge from the other, allowing the features to produce stronger correlations.
In addition, the proposed Bi-PointFlowNet adopts the coarse-to-fine method for multi-scale bidirectional correlation extraction.

We evaluated the proposed model on two challenging benchmarks, the FlyingThings3D \cite{mayer2016large} and KITTI  \cite{menze2018object} datasets, under both occluded and non-occluded conditions. On the FlyingThings3D dataset, Bi-PointFlowNet outperforms all existing methods with more than 44\% and 32\% of estimation error reduction on the non-occluded cases and the occluded cases, respectively. To evaluate generalization performance, we trained the models on the synthetic (FlyingThings3D) dataset and evaluated on the real-world LiDAR scan (KITTI Scene Flow 2015) dataset without fine-tuning. Compared to the existing approaches, the results show that Bi-PointFlowNet achieves improved generality with 44\% and 21\% lower error on the the non-occluded and occluded cases, respectively. Our Bi-PointFlowNet also showed better time efficiency while maintaining high accuracy. 

The key contributions of this paper are summarized as follows:
\begin{itemize}
\setlength{\itemsep}{0pt}
\setlength{\parsep}{0pt}
\setlength{\parskip}{0pt}

\item We are the first to apply the bidirectional learning architecture used for a 3D scene flow estimation task based on point cloud. The model can extract bidirectional correlations that significantly improve flow estimation performance.

\item We propose a decomposed form of the bidirectional layer that optimizes the computation count for accelerated bidirectional correlations extraction. 

\item The proposed model achieves the state-of-the-art performance and generality on the synthetic FlyingThings3D and real-world KITTI benchmarks in both occluded and non-occluded conditions.
\end{itemize}


\section{Related Work}
\subsection{Scene Flow Estimation}

The 3D scene flow, first introduced by \cite{vedula1999three}, represents a dense 3D motion vector field for each point on every surface in the scene. Early dense scene flow estimation approaches \cite{huguet2007variational,valgaerts2010joint,pons2007multi,wedel2008efficient,vcech2011scene,menze2015object,mayer2016large,vogel2013piecewise} used stereo RGB images as an input. With the rapid development of 3D sensors and the emergence of point cloud based networks \cite{qi2017pointnet,wu2019pointconv,qi2017pointnet++}, a line of studies proposed estimating the scene flow using the raw 3D point clouds. 
FlowNet3D \cite{liu2019flownet3d} was the first study that estimated the scene flow from two raw point cloud frames through a deep neural network. However, the performance of FlowNet3D was restricted by its single flow correlation. To address this drawback, Gu et al. proposed HPLFlownet \cite{gu2019hplflownet} that captures multi-scale correlations using a bilateral convolutional layer \cite{jampani2016learning,su2018splatnet}. PointPWC-Net \cite{wu2020pointpwc} further improved the performance and efficiency by hierarchically regressing scene flow in a coarse-to-fine manner. There are several other approaches that leveraged the all-to-all correlation, including FLOT \cite{puy2020flot} that learns the all-to-all correlation by solving an optimal transport problem, and FlowStep3D \cite{kittenplon2021flowstep3d} that iteratively aligns point clouds based on the iterative closest point (ICP) algorithm \cite{chen1992object,besl1992method}. However, learning an all-to-all correlation matrix is computationally inefficient when the input point clouds contain a large number of points.

Our Bi-PointFlowNet is inspired by these point cloud based methods. It also adopts the coarse-to-fine architecture to capture the multi-level correlation and to reduce computational overhead. However, the proposed method is different from the existing models as it utilizes bidirectional learning, which collects the contextual information from both source and target features for more accurate estimation.

\begin{figure}
\centering
\includegraphics[width=\linewidth]{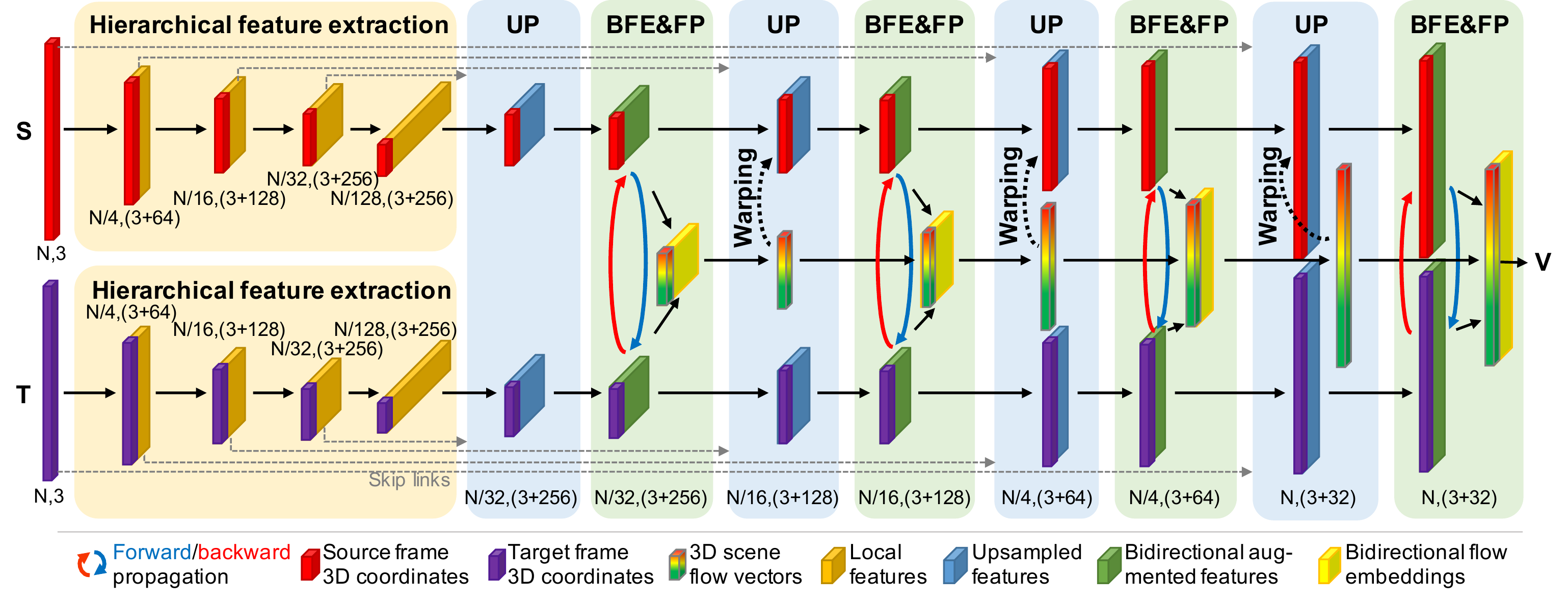}
\caption{Architecture of Bi-PointFlowNet for scene flow estimation. (\textbf{UP} stands for an upsampling layer. \textbf{BFE\&FP} stand for a bidirectional flow embedding layer and a flow prediction layer. They are visualized in the same block for a clear representation.) First, we feed two consecutive input point frames into the shared hierarchical feature extraction module for multilevel feature extraction. Then the upsampling layers propagate features from high levels to low levels and the warping operations are directly applied upon the upsampled points. After each upsampling layer, a bidirectional flow embedding layer is adopted for bidirectional feature (forward feature and backward features) propagation and flow embedding generation. The flow embeddings are immediately fed into the flow prediction layer for scene flow regression according to the current level. The figure is best viewed in color.}
\label{fig:arch}
\end{figure}

\subsection{Bidirectional Models}
The bidirectional models aim to extract features based on both the current and future states. They are able to capture strong contextual information with future knowledge, which is helpful for many time-sequence processing tasks such as natural language processing (NLP). The bidirectional model was first proposed in a bidirectional RNN (BRNN) \cite{schuster1997bidirectional} that learns sequence representations forward and backward through two separate networks. Subsequently, a more powerful structure called bidirectional long short-term memory (BiLSTM) \cite{graves2005framewise} was proposed and successfully applied in frame-wise phoneme classification. Based on these fundamental studies, various approaches \cite{baldi1999exploiting,zhou2016attention,peters2018deep,melamud2016context2vec} have been actively explored. In recent years, the bidirectional encoder representation transformer (BERT) \cite{devlin2018bert} and its variants \cite{liu2019roberta,lan2019albert} have achieved overwhelming performance in various applications including language understanding \cite{yang2019simple,lee2020biobert,wu2019conditional}. 

Recently, a series of studies showed that 2D optical flow estimation can also benefit from bidirectional learning, because optical flow estimation is a type of time series-based task as well. MirrorFlow \cite{hur2017mirrorflow} reused a symmetric optical flow algorithm bidirectionally to extract forward and backward optical flows, which are then constrained by the bidirectional motion and occlusion consistency. Similarly, Wang et al. \cite{wang2018occlusion} also proposed an approach that generates bidirectional optical flows but by reusing a neural network. In addition, Janai et al. \cite{janai2018unsupervised} proposed a method that extracts bidirectional optical flows in a coarse-to-fine manner based on a pyramid structure. Based on the bidirectional models, Hur et al. \cite{hur2019iterative} implemented an architecture that iteratively refines the optical flow estimation by using the previous output.

However, bidirectional learning has not been yet explored in 3D scene flow estimation.
To the best of our knowledge, we propose the first bidirectional model for scene flow estimation based on 3D point cloud. Different from 2D optical flow estimation methods, our proposed model does not reuse an unidirectional flow estimator nor explicitly generate both forward and backward flows. Instead, we only implicitly encode bidirectional features like BRNN and fuse them for only forward flow estimation. Consequently, the model can eliminate redundant computations.

\section{Problem Statement}
Scene flow estimation using a point cloud is to estimate a 3D point-wise motion field in a dynamic scene. The inputs are two consecutive point cloud frames, the source frame $S = \{ p_i = (x_i, f_i) \}_{i=1}^N$ and target frame $T = \{ q_j = (y_j, g_j)\}_{j=1}^M$, where each point consists of a 3D coordinate $x_i, y_j \in \mathbb{R}^3 $ and its corresponding feature $f_i, g_j \in \mathbb{R}^c $. The outputs are 3D motion field vectors $V = \{ v_i \in \mathbb{R}^3 \}_{i=1}^N$ that represent the point-wise non-rigid transformation from the source frame toward the target frame. Our goal is to estimate the best non-rigid transformation $V$ that represents the best alignment from the source frame towards the target frame. Note that $N$ and $M$ denote the number of points in the source and target frame, respectively. However, $N$ and $M$ are not required to be equal because of sparsity and occlusion in a point cloud. Therefore, learning the hard correspondence between the two frames is not feasible. Instead, we directly learn the flow vector for each point in the source frame, as in the most of the recent methods \cite{liu2019flownet3d,wu2020pointpwc,puy2020flot,gu2019hplflownet,kittenplon2021flowstep3d,li2021hcrf}. 

\section{Bi-PointFlowNet}
The proposed Bi-PointFlowNet estimates scene flow using a hierarchical architecture with bidirectional flow embedding extraction. The network accepts two consecutive point cloud frames $S$ and $T$ as an input. The output of the network is the estimated scene flow vectors $V$. As shown in Fig. \ref{fig:arch}, Bi-PointFlowNet consists of four components. First, a hierarchical feature extractor extracts multi-level local features in both input frames. Second, novel bidirectional flow embedding layers are applied at different upsampled levels for multi-level bidirectional correlation extraction. Third, upsampling and warping layers propagate features from higher levels to lower levels. Finally, the flow predictor aggregates bidirectional correlations and propagated features to obtain the flow estimation for each level.

\begin{figure}
\centering
\includegraphics[width=\linewidth]{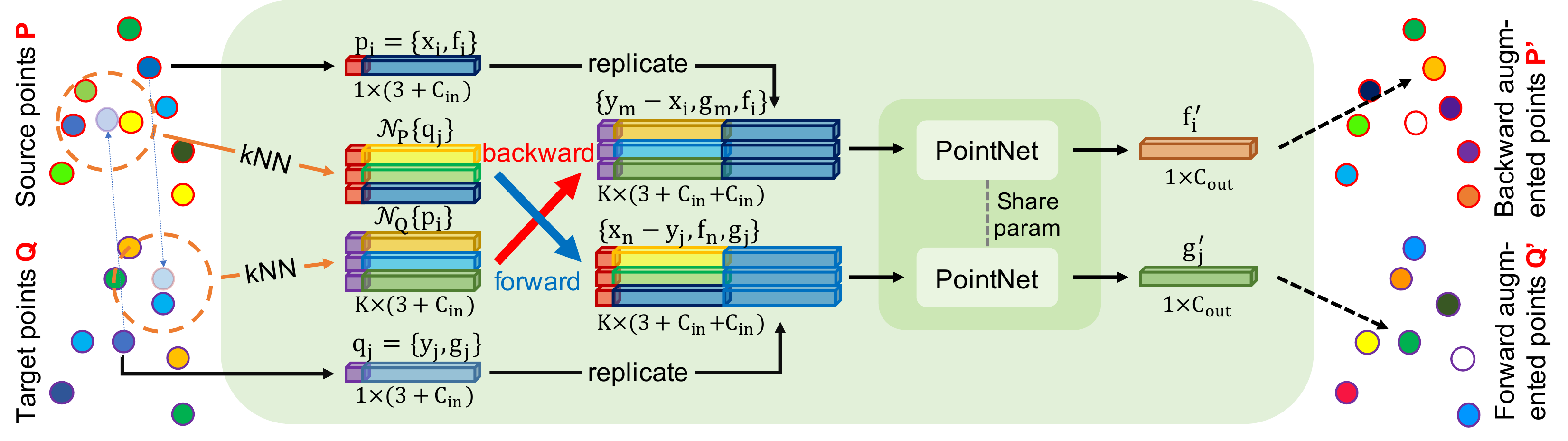}
\caption{Bidirectional feature propagation in the novel Bidirectional Flow Embedding layer. Each point first groups the nearest neighbors from the other point cloud forming a local region. (Forward grouping: a source point groups points from the target points. Backward grouping: a target point groups points from the source points.) Each point in the local region is then concatenated with its own local features propagated from previous feature upsampling. Finally, a PointNet layer with shared parameters accepts the local regions as input and updates the bidirectional augmented features for each point. }
\label{fig:cross}
\end{figure}

\subsection{Hierarchical Feature Extraction}
To extract informative features from point clouds more efficiently and effectively, we adopt the hierarchical feature extraction scheme commonly used in point cloud processing \cite{qi2017pointnet++,wu2019pointconv}. Feature extractions proceed in $L$ levels for generating hierarchical features from dense to sparse. At each level $l$, dense input points and their corresponding features are first subsampled through the \textit{furthest point sampling}, which forms a sparse point set. Then, the \textit{k-nearest neighbor} is used to locally group dense points around every subsampled sparse point. Finally, a \textit{Pointconv} \cite{wu2019pointconv} layer aggregates the features and coordinates from the grouped local points, and generates the local feature for each subsampled point.

\subsection{Bidirectional Flow Embedding}
\label{sec:bfe}

Unlike conventional correlation extraction that uses only unidirectional features between two consecutive frames, we propose a novel bidirectional flow embedding (BFE) layer that provides rich contextual information. The BFE layer first generates bidirectional augmented feature representations through a bidirectional feature propagation (BFP) module, as shown in Fig. \ref{fig:cross}. Then, a conventional \textit{flow embedding} (FE) layer is followed to extract correlation embeddings for flow regression. 

Let the inputs to the BFP module be $P$ and $Q$, where $P \subset S$ and $Q \subset T$ are the subsampled points. 
For each point $p_i \in P$ in the target frame, the BFP module first collects the nearest points from the source frame that forms the group $\mathcal{N}_Q\{p_i\}$. Likewise, the BFP module collects points from the target frame for each point $q_j \in Q$ in the source frame that forms group $\mathcal{N}_P\{q_j\}$. 
Subsequently, the points $p_i$, $q_j$ and their groups $\mathcal{N}_Q\{p_i\}$ , $\mathcal{N}_P\{q_j\}$ are simultaneously processed by a shared PointNet \cite{qi2017pointnet,qi2017pointnet++} layer to generate the bidirectionally-augmented point representations. Thus, the bidirectional-augmented point features, which are backward augmented feature $f'_i$ for $p_i$ and forward augmented feature $g'_j$ for $q_j$, are respectively represented as:

\begin{equation}
f^{'}_i = \mathop{MAX}\limits_{(y_m, g_m) \in \mathcal{N}_Q\{p_i\}} (MLP([y_m - x_i, g_m, f_i])),
\label{eq:forward}
\end{equation}
\begin{equation}
g^{'}_j = \mathop{MAX}\limits_{(x_n, f_n) \in \mathcal{N}_P\{q_j\}} (MLP([x_n - y_j, f_n, g_j])),
\label{eq:backward}
\end{equation}
where MLP and MAX denote the shared MLP and maxpooling layer of the learned PointNet, respectively, and `$[\cdot,\cdot]$' denotes the channel concatenation operator. 

Since the output estimation are only forward directed, a normal unidirectional \textit{flow embedding} (FE) correlation layer captures correlations from the source bidirectionally-augmented points to the target bidirectionally-augmented points after the BFP. 
We name the correlation as the bidirectional flow embedding, because they are extracted from the bidirectional features. 
Note that, the generated augmented points are also fed to the subsequent upsampling layer for hierarchical feature propagation, which will be elaborated in Sec. \ref{subsec:up}.

\subsection{Decomposed Form of Bidirectional Flow Embedding}
\label{subsec:decomp}

The aforementioned BFE layer directly follows the standard procedure (i.e. grouping $\xrightarrow{}$ concatenation $\xrightarrow{}$ MLP $\xrightarrow{}$ max-pooling) to fuse the local information, as presented in \cite{qi2017pointnet++}. However, this procedure requires a large number of operations as it should be executed for each point of the input point cloud. Let the inputs to the BFE module be $P = \{ (x_i, f_i)\in \mathbb{R}^{3+C} \}_{i=1}^{N'}$ and $Q = \{ (y_j, g_j)\in \mathbb{R}^{3+C}\}_{j=1}^{M'}$, and the number of grouping points to be $K$. For convenient analysis, we assume a one-layer MLP whose weights are $W \in \mathbb{R}^{(3+C+C) \times C'}$. Then, $(3+C+C) \times C'$ MLP computations are required for $(N'+ M') \times K$ times. Therefore, the total operation count of BFE is $(N'+ M') \times K \times (3+C+C) \times C'$. However, as the total number of input points is $(N'+ M')$, every $K$ neighbor points are grouped into $(N'+ M')$ groups and then calculated by the MLP. Thus, at least $(N'+ M') \times (K - 1)$ MLP operations are repeatedly calculated.

To optimize this redundancy, we propose a decomposed form of BFE. First, we decompose the MLP weights $W$ into three sub-weights: the weights for local position encoding $W_p \in \mathbb{R}^{3 \times C'}$, the weights for the bidirectional propagated feature $W_b \in \mathbb{R}^{C \times C'}$ and the weights for the replicated feature $W_r \in \mathbb{R}^{C \times C'}$. $W_b$ and $W_r$ are performed at both $P$ and $Q$ before grouping, thus forming transformed features $W_bf_i$, $W_bg_j$, $W_rf_i$ and $W_rg_j$. These transformed features and their corresponding coordinates are then supplied for grouping. Afterwards, only $W_p$ is used for transformation of the grouped local coordinates. Finally, we simply add the transformed local coordinates with the transformed features together and apply the activation function. Therefore, Equation \ref{eq:forward} and \ref{eq:backward} can be transformed as: 

\begin{equation}
f^{'}_i = \mathop{MAX}\limits_{(y_m, W_bg_m) \in \mathcal{N}_Q\{p_i=(x_i, W_rf_i)\}} \sigma(W_p(y_m - x_i)+W_bg_m+W_rf_i),
\end{equation}
\begin{equation}
g^{'}_j = \mathop{MAX}\limits_{(x_n,  W_bf_n) \in \mathcal{N}_P\{q_j=(y_j, W_rg_j)\}} \sigma(W_p(x_n - y_j)+  W_bf_n+W_rg_j),
\end{equation}
where $\sigma$ represents the activation function. Thus, computing $W_b, W_r$ at $P,Q$ only requires $(N'+ M') \times (C+C)\times C'$ operations, while local coordinate transformation requires $(N'+ M') \times (K \times 3) \times C'$ operations. As a result, the total computations count of decomposed BFE is reduced to $(N'+ M') \times (K \times 3 + C+C) \times C'$.

\subsection{Upsampling and Warping}
\label{subsec:up}

The upsampling (UP) layer can propagate features (including flows, local features, and bidirectional augmented points) from sparse levels to dense levels. To reduce the computational cost, we adopt the 3D interpolation using the inverse-distance weighted function based on the $k$ nearest neighbors. 
Let $\{(x^l_j, f^l_j)\}_{j=1}^{N^l}$ denotes the coordinates and features from a high level, and $\{x^{l-1}_i\}_{i=1}^{N^{l-1}}$ denotes the coordinates from a low level through a super link, where $N^{l-1}$ and $N^l$ are the number of points and $N^{l-1} > N^l$. The interpolated feature of a dense point $\{x^l_i\}$ is defined as:

\begin{equation}
f^{l-1}_i = \frac{\sum_{j=1}^k w(x^l_j, x^{l-1}_i)f^l_j}{\sum_{j=1}^k w(x^l_j, x^{l-1}_i)},
\end{equation}
where  $w(x^l_j, x^{l-1}_i) = 1 / ||x^l_j-x^{l-1}_i||_2$, and $k = 3$ by default.

The upsampled scene flows are immediately accumulated to the source frame in order to obtain a frame closer to the corresponding target frame through a warping layer. This process can be simply denoted as $x_w^l = x^l + v^l$ for each source point $x^l$ from the $l$-th level, where $v^l$ denotes an upsampled flow vector. 
Through warping, the warped points gradually become close to the target frame. Thus, the subsequent BFE layer can easily group more valuable points with high semantic similarity that can promote more accurate flow estimation. In addition, accurate flow estimation of the current level also enhances warping at the next level. 

\subsection{Scene Flow Prediction}

We implement a scene flow predictor in order to regress the scene flow vector. For each level, the inputs are the upsampled flows and features from the upsampling layer, and bidirectional flow embeddings from the BFE layer. First, the predictor uses a \textit{Pointconv} to produce smooth features by locally fusing these features and flows around each warped source point. Subsequently, a MLP transforms the smooth high-dimensional features into 3-dimensional scene flow vectors for all points. Since the predictor only focuses on a small region around each warped source point, the outputs from the last MLP layer are point-wise flow residuals, as in \cite{wang2021residual,ouyang2021occlusion}. Afterwards, the residuals are further accumulated with the upsampled flows forming the output flow estimation for the current level.

\subsection{Loss Function}
The training process adopts the multi-level supervision manner used in previous studies for optical flow estimation \cite{dosovitskiy2015flownet,tam2012registration} and scene flow estimation \cite{wang2021hierarchical,wu2020pointpwc}. At each level, the estimated flows are supervised by the L2 loss. Let $\{v_i^l\}_{i=1}^{N^l}$ denote the scene flow vectors estimated from the $l$-th level and $\{\hat{v}_i^l\}_{i=1}^{N^l}$ denote the ground truth scene flow vectors of the $l$-th level. The training loss is defined as: 
\begin{equation}
\mathcal{L} = \sum_{l=0}^{L-1}\alpha^l\sum_{i=1}^{N^l}\|\hat{v}_i^l-v_i^l\|_2,
\end{equation}
where $\alpha^l$ is the weight of the loss function at level $l$. The weights are set to be $\alpha^0 = 0.16$, $\alpha^1 = 0.08$, $\alpha^2 = 0.04$, $\alpha^3 = 0.02$ by default.

\section{Experiments}
\subsection{Experimental Settings}

We conducted experiments on an NVIDIA TITAN RTX GPU with PyTorch. As shown in Fig. \ref{fig:arch}, we implemented a hierarchical model with $L=4$ levels. We used $N=M=8,192$ points as inputs. The numbers of subsampled points of each level  are defined as $N^1=2,048$, $N^2=512$, $N^3=256$, and $N^4=64$. As in the previous methods, we first trained and evaluated networks on the synthetic FlyingThings3D \cite{mayer2016large} dataset (Sec. \ref{sec:ft}). Then, to validate the generalization ability, the trained model was directly evaluated on the real-world KITTI Scene Flow 2015 \cite{menze2018object} dataset without any fine-tuning (Sec. \ref{sec:kitti}).

\subsection{Evaluation Metrics}
For a fair comparison, we adopted the same evaluation metrics as used in recent works \cite{gu2019hplflownet,wu2020pointpwc,puy2020flot,kittenplon2021flowstep3d,li2021hcrf}.

\begin{itemize}
\setlength{\itemsep}{0pt}
\setlength{\parsep}{0pt}
\setlength{\parskip}{0pt}
\item \textbf{EPE3D$_{full}$ (m)}: the main evaluation metric measuring end-point-error $\|\hat{v}_i^l-v_i^l\|_2$ averaged over \textbf{all} point.

\item \textbf{EPE3D (m)}: the main evaluation metric measuring end-point-error $\|\hat{v}_i^l-v_i^l\|_2$ averaged each \textbf{non-occluded} point.

\item \textbf{ACC3DS}: the percentage of points whose EPE3D $<$ 0.05m or relative error $<$ 5\%.

\item \textbf{ACC3DR}: the percentage of points whose EPE3D $<$ 0.1m or relative error $<$ 10\%.

\item \textbf{Outliers3D}: the percentage of points whose EPE3D $>$ 0.3m or relative error $>$ 10\%.

\item \textbf{EPE2D (px)}: 2D end-point-error measured by projecting points back to the 2D image plane, which is a common metric for optical flow evaluation.

\item \textbf{ACC2D}: the percentage of points whose EPE2D $<$ 3px or relative error $<$ 5\%.
\end{itemize}

\begin{table}[h!]
\small
\begin{center}
\begin{tabular}{c|c|cccc|cc}
\toprule[2pt]

\multirow{2}{*}{Dataset} & \multirow{2}{*}{Method} & EPE3D & ACC3D  & ACC3D & Outliers & EPE2D & ACC\\
        &       & (m) $\downarrow$ & S $\uparrow$ & R $\uparrow$ & 3D $\downarrow$ & (px) $\downarrow$ & 2D $\uparrow$ \\
\hline
\multirow{7}{*}{FT3D$_s$}&FlowNet3D \cite{liu2019flownet3d}  
& 0.113     & 0.412     & 0.771     & 0.602 &5.974  &0.569\\ 
&HPLFlowNet \cite{gu2019hplflownet} 
& 0.080     & 0.614     & 0.855     & 0.429 &4.672  &0.676\\ 
&PointPWC  \cite{wu2020pointpwc}  
& 0.059     & 0.738     & 0.928     & 0.342 &3.239  &0.799\\ 
&FLOT   \cite{puy2020flot}     
& 0.052     & 0.732     & 0.927     & 0.357 &-      &-\\ 
&HCRF-Flow \cite{li2021hcrf} 
& 0.048     & 0.835     & 0.950     & 0.261 &2.565  &0.870\\
&FlowStep3D \cite{kittenplon2021flowstep3d} 
& 0.045     & 0.816     & 0.961     & 0.216 &-      &- \\
\cline{2-8}
&\textbf{Ours}        & \textbf{0.028}     &\textbf{0.918}        &\textbf{0.978}      &\textbf{0.143}       &\textbf{1.582} & \textbf{0.929}\\

\hline
\multirow{7}{*}{KITTI$_s$}&FlowNet3D \cite{liu2019flownet3d}  
  & 0.177 & 0.374     & 0.668     & 0.527 &7.214    &0.509\\ 
&HPLFlowNet \cite{gu2019hplflownet} 
   & 0.117 & 0.478     & 0.778     & 0.410 &4.805   &0.593\\ 
&PointPWC  \cite{wu2020pointpwc}  
   & 0.069 & 0.728     & 0.888     & 0.265 &1.902   &0.866\\ 
&FLOT   \cite{puy2020flot}     
   & 0.056 & 0.755     & 0.908     & 0.242 &-   &-\\ 
&HCRF-Flow \cite{li2021hcrf} 
   & 0.053 & 0.863     & 0.944     & 0.179 &2.070  &0.865\\
&FlowStep3D \cite{kittenplon2021flowstep3d} 
   & 0.054 & 0.805     & 0.925      & 0.149 &-   &-\\
\cline{2-8}
&\textbf{Ours}           &\textbf{0.030}  &\textbf{0.920}      &\textbf{0.960}      &\textbf{0.141} &\textbf{1.056} & \textbf{0.949}\\
\bottomrule[2pt]

\end{tabular}
\end{center}
\caption{Comparison of the proposed method with previous state-of-the-art methods on the non-occluded FT3D$_s$ and KITTI$_s$ datasets. All methods are trained only on the FT3D$_s$ dataset.}
\label{tab:sota_s}
\end{table}

\begin{table}[h!]
\small
\begin{center}
\begin{tabular}{c|c|c|cccc}
\toprule[2pt]

\multirow{2}{*}{Dataset} & \multirow{2}{*}{Method} & EPE3D$_{full}$& EPE3D & ACC3D  & ACC3D & Outliers \\
        &       & (m) $\downarrow$ & (m) $\downarrow$ & S $\uparrow$ & R $\uparrow$ & 3D $\downarrow$ \\
\hline
\multirow{6}{*}{FT3D$_o$}&FlowNet3D \cite{liu2019flownet3d}  
& 0.211     & 0.157     & 0.228    & 0.582  & 0.804\\ 
&HPLFlowNet \cite{gu2019hplflownet} 
& 0.201     & 0.168     & 0.262    & 0.574  & 0.812\\ 
&FLOT   \cite{puy2020flot}     
& 0.250     & 0.153     & 0.396    & 0.660 & 0.662 \\ 
&PointPWC  \cite{wu2020pointpwc}  
& 0.195     & 0.155     & 0.416    & 0.699 & 0.638 \\ 
&OGSFNet \cite{ouyang2021occlusion}     
& 0.163     & 0.121     & 0.551    & 0.776 & 0.518 \\ 
\cline{2-7}
&RAFT-3D (16 iters) \cite{teed2021raft}     
& -     & \textbf{0.064}      & \textbf{0.837}    & 0.892 & - \\ 

\cline{2-7}
&\textbf{Ours}        & \textbf{0.102}     &0.073    &0.791     &\textbf{0.896}      &\textbf{0.274}       \\

\hline
\multirow{7}{*}{KITTI$_o$}&FlowNet3D \cite{liu2019flownet3d}  
  & 0.183 & - & 0.098     & 0.394     & 0.799 \\ 
&HPLFlowNet \cite{gu2019hplflownet} 
   & 0.343 & -  & 0.103     & 0.386     & 0.814 \\ 
&FLOT   \cite{puy2020flot}     
   & 0.130 & -  & 0.278     & 0.667     & 0.529 \\ 
&PointPWC  \cite{wu2020pointpwc}  
   & 0.118 & -  & 0.403     & 0.757     & 0.496 \\ 
&OGSFNet \cite{ouyang2021occlusion}     
& 0.075   & -   & 0.706     & 0.869    & 0.327 \\ 
\cline{2-7}
&\textbf{Ours}           &\textbf{0.065}  & -  &\textbf{0.769}      &\textbf{0.906}      &\textbf{0.264} \\

\bottomrule[2pt]

\end{tabular}
\end{center}
\caption{Comparison of the proposed method with previous state-of-the-art methods on the occluded FT3D$_o$ and KITTI$_o$ datasets. All methods are trained only on the FT3D$_o$ dataset.}
\label{tab:sota_o}
\end{table}

\subsection{Training and Evaluation on FlyingThings3D}
\label{sec:ft}

FlyingThing3D \cite{mayer2016large} is a synthetic dataset composed of 19,640 pairs of frames for training and 3,824 pairs of frames for testing. Each frame consists of stereo and RGB-D images rendered from scenes with multiple moving objects sampled from ShapeNet \cite{chang2015shapenet} dataset. We trained and evaluated our proposed model based on two versions of datasets prepared by different pre-processing methodologies. The first version is FT3D$_s$, which removes the occluded points after transforming the image data into points, as suggested in \cite{gu2019hplflownet,wu2020pointpwc,puy2020flot,kittenplon2021flowstep3d}. The second version, FT3D$_o$ introduced by \cite{liu2019flownet3d,puy2020flot,ouyang2021occlusion}, remains the occluded points. 
The input points of $N=8,192$ are randomly sampled from each frame with non-correspondence.

For training, we used the Adam optimizer \cite{kingma2014adam} with beta1 = 0.9, beta2 = 0.999, and starting learning rate $\alpha$ = 0.0001. The learning rate is reduced by half every 80 epochs. We trained the model for 560 epochs.

\noindent
\textbf{Results.} We report the performance of the proposed model compared to other state-of-the-art approaches \cite{liu2019flownet3d,gu2019hplflownet,wu2020pointpwc,puy2020flot,kittenplon2021flowstep3d,ouyang2021occlusion}. On the non-occluded FlyingThings3D dataset, the proposed Bi-PointFlowNet achieved a new state-of-the-art record on all evaluation metrics based on point cloud, as shown in Table \ref{tab:sota_s}. It outperformed all recent state-of-the-art methods with more than 44\% reduction of estimation error. When compared to the similar coarse-to-fine PointPWC \cite{wu2020pointpwc}, our model achieved an error reduction of 52\%. On the other hand, Table \ref{tab:sota_o} also shows remarkable performance of our work when handling occluded data. Our Bi-PointFlowNet improved the state-of-the-art performance by 32\%. In addition, we also compared our method with RGB-D image-based RAFT-3D \cite{teed2021raft}. Table \ref{tab:sota_o} shows that our method achieved comparable performance than Raft-3D with 16 iterations. Although our method did not achieve better EPE3D and ACC3DS, it outperformed Raft-3D for the ACC3DR metric. Despite a minor increase in errors, we report that our model requires significant less computation (13.3GFLOPs) and parameter size (7.9M) than RAFT-3D (329GFLOPs, 45M), making it more applicable to time-sensitive low-power applications. According to \cite{teed2021raft}, we expect that RAFT-3D having similar computation to ours with less iteration will yield much worse accuracy than ours.

\subsection{Generalization on KITTI}
\label{sec:kitti}

In order to evaluate the generalization ability of Bi-PointFlowNet to real-world data, we followed the same evaluation strategy as in the recent studies \cite{liu2019flownet3d,gu2019hplflownet,wu2020pointpwc,puy2020flot,kittenplon2021flowstep3d,ouyang2021occlusion}. We directly tested the trained model on the real-world KITTI \cite{menze2018object} dataset without fine-tuning. The KITTI dataset contains 200 training and 200 testing sets. However, due to unprovided disparities in the testing set and in parts of the training set, we used 142 scenes (non-occluded) and 150 scenes (occluded) from the training set with available raw point clouds. For fair comparison of our method with the previous approaches \cite{liu2019flownet3d,gu2019hplflownet,wu2020pointpwc,puy2020flot,kittenplon2021flowstep3d,ouyang2021occlusion}, we followed the common step that removes ground points by height $<$ 0.3 m. According to the preparation of the FlyingThings3D dataset, both the non-occluded KITTI$_s$ and occluded KITTI$_o$ datasets are created.
\begin{figure}
\includegraphics[width=\linewidth]{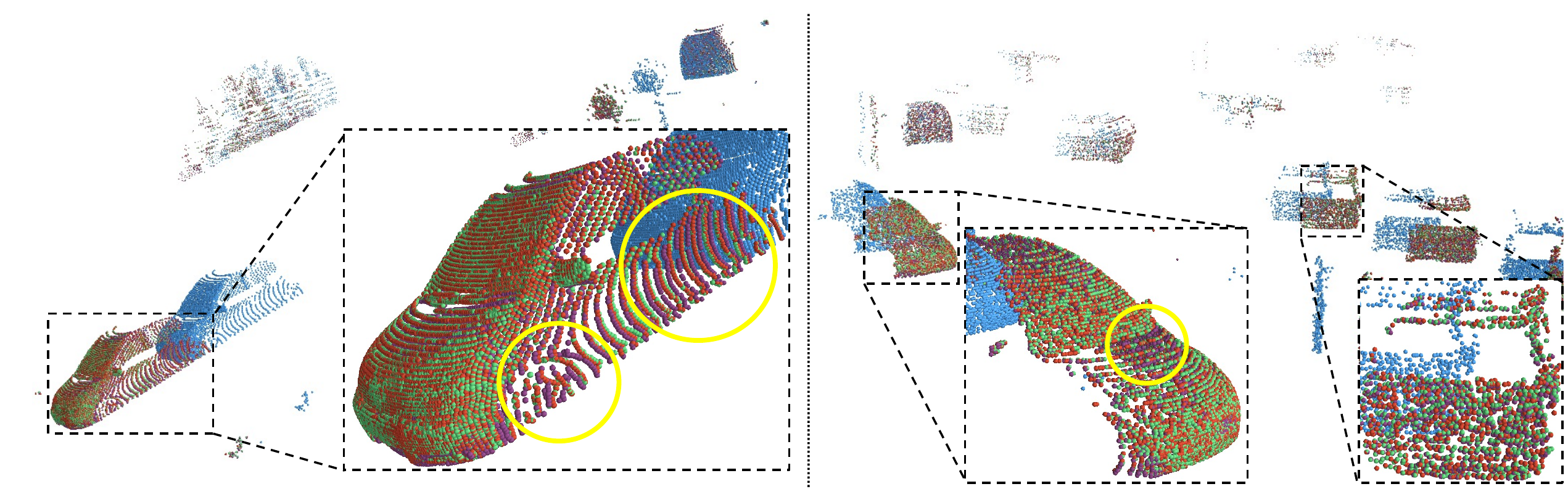}
\caption{Qualitative results of Bi-PointFlowNet on the non-occluded KITTI$_s$ dataset. Points are colored to indicate points as from \textcolor{cyan}{source frame}, \textcolor{red}{target frame}, \textcolor[RGB]{150,52,132}{unidirectional PointPWCNet estimated points} (source frame + scene flow) or as \textcolor[RGB]{40,150,90}{bidirectional Bi-PointFlowNet estimated points} (source frame + scene flow). }
\label{fig:kitti}
\end{figure}

\noindent
\textbf{Results.}
The generalization results on KITTI$_s$ and KITTI$_o$ are listed in Table \ref{tab:sota_s} and \ref{tab:sota_o}, respectively. Our method significantly outperforms other methods on all metrics by a large margin. Table \ref{tab:sota_s} represents that the model outperforms the previous state-of-the-art method by 44\% on the main EPE3D metric. Compared with previous coarse-to-fine network, PointPWC-Net \cite{wu2020pointpwc}, our method achieves 56\% error reduction. Meanwhile, Table \ref{tab:sota_o} shows that our model outperforms the previous state-of-the-art method by 21\% of error reduction.  In addition, we present the qualitative results on the non-occluded cases of the KITTI$_s$ dataset in Fig. \ref{fig:kitti}. The results show that our Bi-PointFlowNet reduced the estimation error for all points compared to the unidirectional coarse-to-fine PointPWC-Net. Furthermore, Bi-PointFlowNet is able to keep more accurate surface and contour details than PointPWC-Net (marked in the yellow circles in Fig. \ref{fig:kitti}).

\subsection{Ablation Study}
\noindent
\textbf{Ablation of the bidirectional flow embedding layer.}
As described in Sec. \ref{sec:bfe}, the key component of the proposed bidirectional flow embedding layer is the bidirectional feature propagation module, which is followed by the conventional unidirectional flow embedding layer. To evaluate the contribution of BFP, we implemented an ablation model that removes the BFP module resulting in a unidirectional network. We compare the performance of this ablation model with our proposed full model in Table \ref{tab:abla}. The results show that the proposed BFP module significantly improved the performance on all metrics with a large margin. Especially, the EPE3D error of the generality test on the KITTI$_s$ dataset was reduced by 43\%, which shows important implications in the real-world applications. In addition, the ablation model without BFP and original PointPWC-Net are both coarse-to-fine architectures. However, due to the introduction of the residual in the flow predictor, the ablation model still outperformed PointPWCNet, according to Table \ref{tab:sota_s} and \ref{tab:abla}.

\begin{table}[t!]
\small
\begin{center}
\begin{tabular}{c|c|cccc|cc}
\toprule[2pt]

\multirow{2}{*}{Dataset} &  \multirow{2}{*}{BFP} & EPE3D & ACC3D  & ACC3D & Outliers& EPE2D  & ACC\\
        &    & (m) $\downarrow$  & S $\uparrow$  & R $\uparrow$ & 3D $\downarrow$ & (px) $\downarrow$ & 2D $\uparrow$\\ 
\hline
\multirow{2}{*}{FT3D$_s$}
&  $\times$
& 0.042     & 0.836     & 0.962     & 0.263 &2.270  &0.882\\ 
& $\surd$       & \textbf{0.028}     &\textbf{0.918}        &\textbf{0.978}      &\textbf{0.143}       &\textbf{1.582} & \textbf{0.929}\\

\hline
\multirow{2}{*}{KITTI$_s$}
& $\times$
    & 0.053 & 0.858     & 0.930    & 0.194 &1.894    &0.880\\ 
& $\surd$           &\textbf{0.030}  &\textbf{0.920}      &\textbf{0.960}      &\textbf{0.141} &\textbf{1.056} & \textbf{0.949}\\
\bottomrule[2pt]

\end{tabular}
\end{center}
\caption{Ablation of the bidirectional flow embedding layer. BFP indicates whether the BFP module is used. All methods are trained only on the FlyingThings3D dataset.}
\label{tab:abla}
\end{table}

\begin{table}[t!]
\small
\begin{center}
\begin{tabular}{c|c|cccc|cc|cc}
\toprule[2pt]

\multirow{2}{*}{Dataset} & \multirow{2}{*}{Decomp.} & EPE3D & ACC3D  & ACC3D & Outliers& EPE2D  & ACC & \multirow{2}{*}{GFLOPs} & Runtime\\
        & & (m) $\downarrow$  & S $\uparrow$  & R $\uparrow$ & 3D $\downarrow$ & (px) $\downarrow$ & 2D $\uparrow$ & & (ms)\\ 
\hline
\multirow{2}{*}{FT3D$_s$} 
& $\times$ & 0.029   & 0.917     & 0.977     & \textbf{0.142} &1.633  &0.928 & 23.8 & 61.2\\ 
& $\surd$ & \textbf{0.028}     &\textbf{0.918}        &\textbf{0.978}      &0.143       &\textbf{1.582} & \textbf{0.929}& 13.3 & 40.5\\ 
\hline
\multirow{2}{*}{KITTI$_s$} 
& $\times$ & 0.030   & \textbf{0.925}     & \textbf{0.965}     & \textbf{0.133} &1.079  &\textbf{0.951}& 23.8 &61.2\\
& $\surd$  &\textbf{0.030}  &0.920      &0.960      &0.141 &\textbf{1.056} & 0.949& 13.3 &40.5\\ 
\bottomrule[2pt]

\end{tabular}
\end{center}
\caption{Ablation of the decomposed form of the bidirectional flow embedding layer. Decomp. indicates whether using decomposed form of BFE. GFLOPs indicates the total operation count. All methods are trained only on the FT3D$_s$ dataset.}
\label{tab:decomp}
\end{table}

\begin{table}[t!]
\small
\begin{center}
\begin{tabular}{c|ccc}
\toprule[1.5pt]
\multirow{2}{*}{Model} & FT3D & KITTI & Param  \\
 & EPE3D (m) & EPE3D (m) & size (M) \\
\hline
PointPWC & 0.059 & 0.069 & 7.72M\\
PointPWC + BFP & \textbf{0.051} & \textbf{0.059} & 7.98M\\
\hline
FlowNet3D & 0.157 & 0.173 & 1.23M\\
Deeper FlowNet3D & 0.160 &0.197 & 1.33M \\
FlowNet3D + BFP & \textbf{0.138} &\textbf{0.118}& 1.33M \\
\bottomrule[1.5pt]

\end{tabular}
\end{center}
\caption{Comparison of the bidirectional feature propagation on PointPWC and FlowNet3D. Although the selected baselines showed strong performance, our proposed BFP still reduced the errors by large margins.}
\label{tab:bfe}
\end{table}

\noindent
\textbf{Ablation of the decomposed form for BFE.} We performed two comparative experiments to evaluate the effectiveness and efficiency of the proposed decomposed form of BFE. The one is Bi-PointFlowNet with original BFE (Sec. \ref{sec:bfe}) and the other one is the model with the decomposed BFE (Sec. \ref{subsec:decomp}). Table \ref{tab:decomp} shows that the model using the decomposed form significantly reduces the total operation count by 44\% and accelerates the inference by 33\% while maintaining the accuracy, compared with the original model. 

\noindent
\textbf{Ablation of our contribution to FlowNet3D and PointPWC.}
We validate the contribution of the proposed bidirectional learning method by applying the BFP module into other state-of-the-art methods, FlowNet3D \cite{liu2019flownet3d} and PointPWC \cite{wu2020pointpwc}. We built two models by directly inserting BFPs before their flow correlation modules. Since adding BFP requires additional parameters, we also implemented a deeper FlowNet3D network with an equivalent amount of parameters as the model with BFP. Please note that, the experiments related to FlowNet3D were evaluated on the occluded datasets while the PointPWC-based experiments were tested on the non-occluded datasets. Table \ref{tab:bfe} indicates that the proposed BFP achieves the excellent efficiency and effectiveness. With 0.2M (only 3\% of total) additional parameters to the PointPWC, the performance is improved with a 13\% error reduction. Moreover, the combination of FlowNet3D and BFP significantly reduce the generalization error by 31\%. Furthermore, the ablation of the deeper FlowNet3D reveals the improved performance is owing to the bidirectional strategy rather than an effect of increased number of parameters.

\subsection{Runtime}
\label{sec:runtime}
We compare the running time of our proposed methods to that of other state-of-the-art approaches in Table \ref{tab:time}. We measured the runtimes of all methods on a single NVIDIA TITAN RTX GPU. The model ran in 40.5 ms, which is faster than the coarse-to-fine PointPWC \cite{wu2020pointpwc} due to the use of the BFE decomposition.
Moreover, compared with other recent advanced approaches \cite{puy2020flot,kittenplon2021flowstep3d}, our methods outperformed by a large margin in terms of running time while achieving superior accuracy and generality.

\begin{table}[t!]
\small
\begin{center}
\begin{tabular}{c|c|c|c|c}
\toprule[1.5pt]
Method & PointPWC \cite{wu2020pointpwc}&  FLOT \cite{puy2020flot} & FlowStep3D \cite{kittenplon2021flowstep3d} & Ours \\
\hline
 Runtime (ms)  & 51.3 & 289.6 & 820.8 & 40.5 \\
\bottomrule[1.5pt]
\end{tabular}
\end{center}
\caption{Runtime comparison. The results are evaluated on a single TITAN RTX GPU. }
\label{tab:time}
\end{table}

\section{Conclusion}
We presented Bi-PointFlowNet for accurate and fast scene flow estimation. Our proposed network leverages a novel bidirectional flow embedding module that worked with hierarchical feature extraction and propagation to accurately estimate flow. For further accelerating inference, the proposed method applied the decomposed form of the bidirectional flow embedding layer that removes the redundant computations. Experimental results on two challenging datasets showed our network significantly outperformed previous state-of-the-art methods under both non-occluded and occluded conditions. The proposed models also demonstrated excellent time efficiency, allowing the models to be further applied to resource-limited devices, such as wearable devices, drones, IoT edge devices, etc. \\

\noindent
\textbf{Acknowledgement.} This work was partly supported by the National Research Foundation (NRF) grants (2022R1F1A1074142, 2022R1A4A3032913) and Institute of Information and Communication Technology Planning \& Evaluation (IITP) grants (IITP-2019-0-00421, IITP-2020-0-00821, IITP-2021-0-02052, IITP-2021-0-02068), funded by the MSIT (Ministry of Science and ICT) of Korea. Wencan Cheng was partly supported by the China Scholarship Council (CSC).

%
%

\end{document}